\documentclass[runningheads]{llncs}

\usepackage{eccv}

\usepackage{eccvabbrv}

\usepackage{graphicx}
\usepackage{booktabs}

\usepackage[accsupp]{axessibility}  

\usepackage{hyperref}

\usepackage{orcidlink}

\usepackage{mathtools}

\usepackage{hyperref}
\usepackage{url}
\usepackage{wrapfig}
\usepackage{etoolbox}
\BeforeBeginEnvironment{wrapfigure}{\setlength{\intextsep}{0pt}}
\usepackage{caption}
\usepackage{subcaption}
\usepackage{multirow}
\usepackage{pifont}
\usepackage{minitoc}

\newcommand{\upx}[1]{\textcolor{violet}{\footnotesize \ $\uparrow${#1}}}

\newcommand{\cmark}{\ding{51}}
\newcommand{\xmark}{\ding{55}}

\usepackage{graphicx}
\usepackage{booktabs}
\usepackage{color}
\usepackage{colortbl}  
\usepackage{xcolor}
\usepackage[table]{xcolor}
\usepackage{array}
\usepackage{bbding}
\usepackage{algorithm}
\usepackage{algpseudocode}
\usepackage{ulem}
\usepackage{tabularx}
\usepackage{bbding}
\usepackage{makecell}

\begin{document}

\title{StruVis: Enhancing Reasoning-based Text-to-Image Generation via Thinking with Structured Vision} 

\titlerunning{StruVis}

\author{
Yuanhuiyi Lyu\inst{1}\thanks{Equal contribution.}
\and
Kaiyu Lei\inst{1}\protect\footnotemark[1]
\and
Ziqiao Weng\inst{1}
\and
Xu Zheng\inst{1}
\and
Lutao Jiang\inst{1}
\and
Teng Li\inst{4}
\and
Yangfu Li\inst{5}
\and
Ziyuan Huang\inst{2}
\and
Linfeng Zhang\inst{3}
\and
Xuming Hu\inst{1,4}\thanks{Corresponding author}
}

\authorrunning{Y. Lyu et al.}

\institute{
Hong Kong University of Science and Technology (Guangzhou) 
\and
Ant Group
\and
Shanghai Jiao Tong University
\and
Hong Kong University of Science and Technology
\and
East China Normal University\\
\email{ryan.lyu.mail@gmail.com, leikaiyu367@gmail.com, xuminghu@hkust-gz.edu.cn}
\\
}

\maketitle
\begin{figure}[h!]
    \centering
    \vspace{-16pt}
    \includegraphics[width=\textwidth]{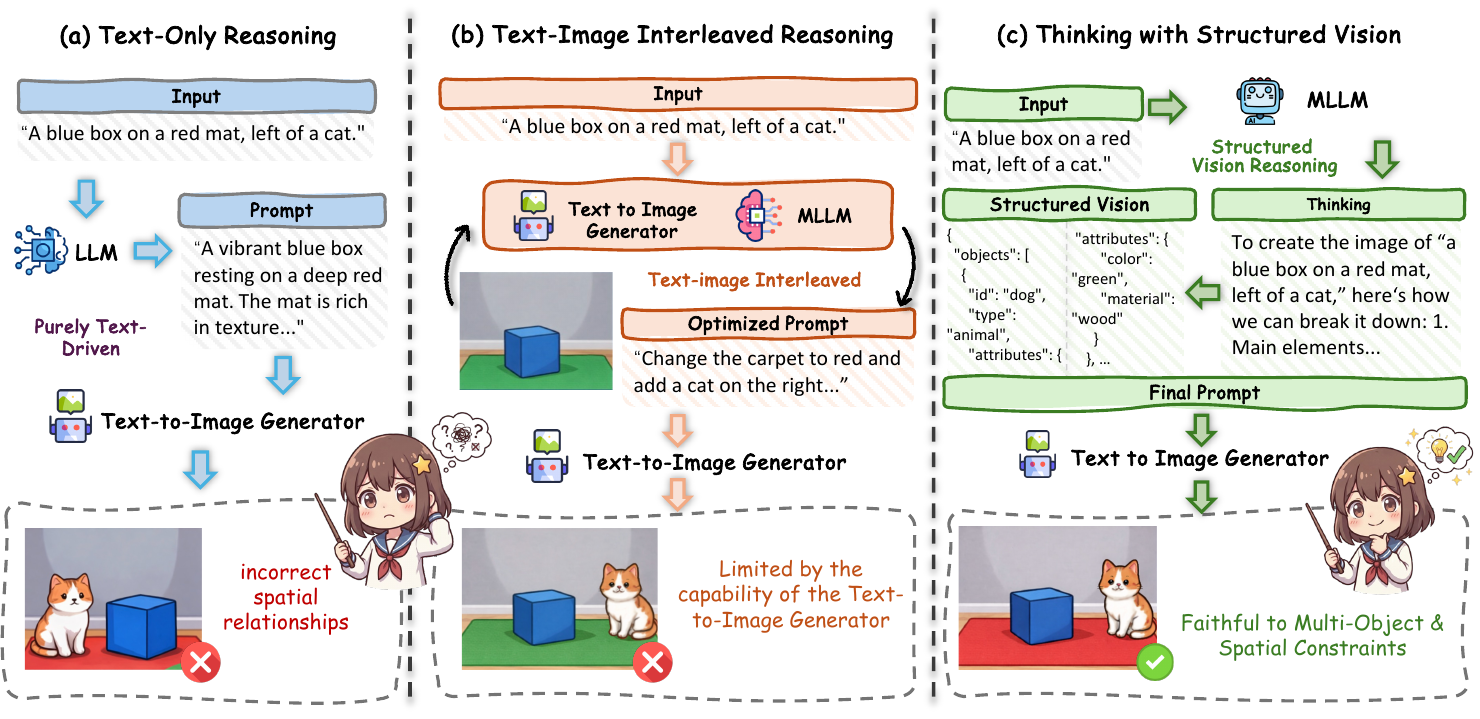}
    \vspace{-12pt}
    \caption{The overall of \textbf{(a)} Text-Only Reasoning, \textbf{(b)} Text-Image Interleaved Reasoning, and \textbf{(c)} Our Thinking with Structured Vision.}
    \vspace{-28pt}
    \label{fig: teaser}
\end{figure}

\begin{abstract}
Reasoning-based text-to-image (T2I) generation requires models to interpret complex prompts accurately. Existing reasoning frameworks can be broadly categorized into two types: \textbf{(1)} Text-Only Reasoning, which is computationally efficient but lacks access to visual context, often resulting in the omission of critical spatial and visual elements; and \textbf{(2)} Text–Image Interleaved Reasoning, which leverages a T2I generator to provide visual references during the reasoning process. While this approach enhances visual grounding, it incurs substantial computational costs and constrains the reasoning capacity of MLLMs to the representational limitations of the generator.
To this end, we propose \textit{StruVis}, a novel framework that enhances T2I generation through \textbf{Thinking with Structured Vision}. Instead of relying on intermediate image generation, StruVis employs text-based structured visual representations as intermediate reasoning states, thereby enabling the MLLM to effectively \textit{``perceive''} visual structure within a purely text-based reasoning process. Powered by this, the reasoning potential for T2I generation of the MLLM is unlocked through \textit{structured-vision-guided reasoning}.
Additionally, as a generator-agnostic reasoning framework, our proposed StruVis can be seamlessly integrated with diverse T2I generators and efficiently enhance their performance in the reasoning-based T2I generation.
Extensive experiments demonstrate that StruVis achieves significant performance improvements on the reasoning-based T2I benchmarks, \eg, a \textbf{4.61\%} gain on the T2I-ReasonBench and a \textbf{4\%} gain on the WISE.

\end{abstract}


\section{Introduction}
Text-to-image generation has advanced significantly in recent years. However, as real-world prompts become more complex, they increasingly demand more than the mere rendering of a single object in a specific style. Such prompts often specify multi-object compositions with explicit constraints, including object counts, attribute bindings, spatial relationships, and global layout requirements. Satisfying these constraints necessitates reasoning-based T2I generation, wherein the model must accurately interpret intricate instructions, devise a coherent scene plan, and faithfully translate that plan into a visually consistent structure.
Existing reasoning frameworks for reasoning-based T2I generation can be broadly categorized into two approaches: \textbf{(1)} Text-Only Reasoning, which performs multi-step reasoning solely in the text domain, typically generating an optimized prompt or structured plan for the generator; and \textbf{(2)} Text–image interleaved reasoning, which alternates between reasoning and intermediate image generation, using the generated images as visual references to guide subsequent steps. While both paradigms have demonstrated promise, each faces fundamental limitations that hinder accurate and efficient constraint satisfaction.

As shown in Fig.~\ref{fig: teaser} \textbf{(a)}, Text-Only Reasoning methods typically utilize the MLLM as a planner to decompose a complex prompt into sub-goals, \eg, object inventories, attribute assignments, relation constraints, and ordering, before rewriting or refining the final prompt for the T2I generative model. Since no intermediate images are integrated into the reasoning, this framework is computationally efficient and easily integrated with various generators. 
However, the Text-Only Reasoning that relies solely on the text prompt lacks the necessary visual context, which leads to the omission of essential visual elements in the reasoning process. As a result, some generated images lack visual details and contain incorrect spatial relationships between objects.

Text–Image Interleaved Reasoning incorporates visual context into the reasoning. As shown in Fig.~\ref{fig: teaser} \textbf{(b)}, the model generates intermediate images during multi-step reasoning and subsequently inspects them to correct errors and refine subsequent steps. This approach substantially enhances visual quality, as the reasoning process can directly align decisions with what the generator actually generated. However, interleaving introduces two limitations. First, the repeated T2I calls lead to high latency and increased costs. Second, the interleaved reasoning framework constrains the reasoning capacity of MLLMs by limiting it to the representational capabilities of the image generator. For example, during the reasoning process, if the image generator fails to generate intermediate images that follow the instructions, the reasoning process of the MLLM will be disrupted. This causes the performance of the Text-Image Interleaved Reasoning framework to be limited by the capabilities of the image generator.

In this work, we propose \textit{StruVis}, a novel framework for reasoning-based T2I generation via \textit{\textbf{thinking with structured vision}}. Instead of relying on intermediate image generation, StruVis introduces text-based structured visual representations as intermediate reasoning states. 
\emph{The \textbf{\textit{core idea}} is to enhance the reasoning-based T2I generation via text-based structured visual representation.}
Specifically, we first construct a \textit{``Text-based Structured Vision''} interleaved CoT dataset (StruVis-CoT), which incorporates vision into the CoT data using structured text. We then train the MLLM with SFT on this dataset to adapt the MLLM to this CoT reasoning format. Finally, we perform GRPO training to further align the MLLM with the ``Thinking with Structured Vision'' reasoning approach, enabling the MLLM to ``see'' intermediate state images in a structured visual representation during the prompt reasoning for T2I generation.

StruVis is model-agnostic and can be seamlessly integrated with a wide range of base models, offering an efficient reasoning enhancement for reasoning-based T2I generation. Extensive experiments on reasoning-based T2I generation benchmarks demonstrate that StruVis consistently delivers significant improvements for reasoning-based T2I generation, \emph{including a \textbf{4.61\%} gain on T2I-ReasonBench and a \textbf{4\%} gain on WISE.} These results highlight both its effectiveness and practicality as an efficient reasoning framework for reasoning-based T2I generation.
\textbf{Our key contributions are summarized as follows}:
\begin{itemize}
    \item We introduce \textbf{StruVis}, a novel framework for reasoning-based T2I generation, which enhances reasoning via \textbf{\emph{structured visual representation}}.
    \item We construct the \textbf{\emph{StruVis-CoT}} dataset, which incorporates visual context into the Chain-of-Thought data using text-based structured vision.
    \item We present extensive experimental results on T2I reasoning benchmarks, demonstrating significant improvements with our proposed StruVis, including a \emph{\textbf{4.61\% gain} on T2I-ReasonBench and a \textbf{4\% gain} on WISE.}
\end{itemize}

\section{Related Work}
\subsection{RL-based Reasoning}

Reinforcement learning training methods, \eg, Group Relative Policy Optimization (GRPO) \cite{shao2024deepseekmath}, have become a dominant approach for training large language models (LLMs) for reasoning tasks. GRPO and similar RL algorithms offer several advantages in this context. By framing reasoning as a sequential decision-making process, they allow for dynamic adjustments to model behavior based on rewards, facilitating complex problem-solving~\cite{guo2025deepseek,yang2025qwen3,yang2025qwen25,fan2025grit,he2025skywork,xie2025logic,yu2025dapo,liu2025drgrpo,zhang2025critiquegrpo}. The ability to directly optimize policies using reward signals has led to improved generalization and higher efficiency in tasks requiring nuanced inference and understanding.
Building upon the success of RL in text-based reasoning, multimodal RL training has also seen significant advancements, particularly in tasks involving both text and images~\cite{zhou2025reinforced,xue2025dancegrpo,liu2025flow}. One prominent example of this success is the \textit{``Think with Image"} framework \cite{liu2025seg,fan2025grit,ni2025point,wu2025vtool,wang2025pixel,bai2025univg,huang2025visualtoolagent,zhu2025active,yang2025visionthink,su2025openthinkimg,zheng2025deepeyes}, which leverages RL to align textual and visual reasoning for multimodal tasks. In this paradigm, models not only process language but also integrate visual cues to enhance reasoning capabilities, thereby achieving substantial improvements in tasks that require both vision and language understanding. For instance, Thyme~\cite{zhang2025thyme} proposed a novel approach to enhancing multimodal large language models (MLLMs) with autonomous image-manipulation and computational capabilities via reinforcement learning. Thyme significantly advances traditional Think with Image models by making the reasoning process more interactive and flexible via the proposed GRPO-ATS.

\subsection{Reasoning-based T2I Generation}
Text-to-Image (T2I) generation has rapidly advanced, with the goal of producing realistic images based on textual descriptions~\cite{flux,wu2025qwen,cao2025hunyuanimage,deng2025emerging}. However, generating accurate and meaningful images from text remains a challenging task, particularly when complex reasoning is required to bridge the gap between linguistic input and visual output. Reasoning-based T2I generation addresses this challenge by incorporating logical and spatial reasoning into the generative process, ensuring that the generated image aligns not only with the textual description but also with the underlying contextual knowledge and spatial relationships.

Recent works in reasoning-based T2I generation have been inspired by advances in MLLMs, where reasoning capabilities are increasingly leveraged for various tasks. These approaches can be broadly categorized into two main paradigms: 
(1) Text-only Reasoning~\cite{jiang2025t2i,zhang2025reasongen,wang2025pref,liao2025imagegen,lin2025decot,gu2025improving}: This approach focuses on reasoning solely within the textual domain, using sophisticated language models to infer spatial relationships, object interactions, and other contextual elements from the text. While computationally efficient, this method suffers from a lack of access to visual context, which often leads to the omission of critical spatial and visual elements in the generated image. 
(2) Text-Image Interleaved Reasoning~\cite{ye2025visual,qin2025uni,lyu2025understanding,gu2025thinkmorph,zhang2025layercraft,huang2025interleaving,jiang2025draco,liu2025cot,guo2025can,li2025visual}: To overcome the limitations of Text-Only Reasoning, this approach integrates a T2I generator into the reasoning process. By leveraging the generator's visual capabilities, models can better ground their reasoning in visual context. However, this method comes with high computational costs due to the need for continuous image generation during reasoning. Moreover, the reasoning capacity of MLLMs is constrained by the representational limitations of the generator, which may not always capture the full range of visual details required for complex image generation.

To address the challenges, we propose StruVis, a novel framework that enhances T2I generation by introducing a structured vision approach to reasoning. Unlike traditional methods that rely on intermediate image generation, StruVis adopts structured visual representations as intermediate reasoning states. These representations encapsulate visual information in a purely text-based framework, allowing the model to “perceive” visual structure without needing to generate images at each step of the reasoning process. By leveraging structured visual representations, StruVis enables the MLLM to maintain a rich, structured understanding of the vision while reasoning over text, effectively combining the strengths of both visual and text reasoning.

\section{Methodology}

\subsection{Problem Formulation}

Reasoning-based T2I generation aims to generate an image based on complex textual prompts, where the prompts often involve multi-object compositions with explicit constraints such as object counts, spatial relationships, and attributes. Given a user prompt \( P \), the objective is to generate a corresponding image \( I \) that satisfies these constraints. The problem can be formulated as:
\begin{equation}
    I = \text{Generator}(P, \mathcal{C}),
\end{equation}
where \( \mathcal{C} \) represents a set of constraints in the prompt \( P \). 
The generation process involves reasoning over these constraints, interpreting the intricate instructions, and generating an image \( I \) that aligns with the specified constraints. 

The task requires reasoning across multiple steps to decompose the user prompt and map it to an image representation. For reasoning-based T2I generation, this process involves solving the following two main components:
\begin{equation}
\mathcal{R}(P, \mathcal{C}) \rightarrow \{r_1, r_2, \dots, r_n\},
\end{equation}
where each reasoning step \( r_i \) progressively refines the understanding of the prompt until the final image \( I \) can be generated. The goal is to generate a visual representation that respects the constraints while being visually consistent.

\begin{figure}[t!]
    \centering
    \includegraphics[width=\textwidth]{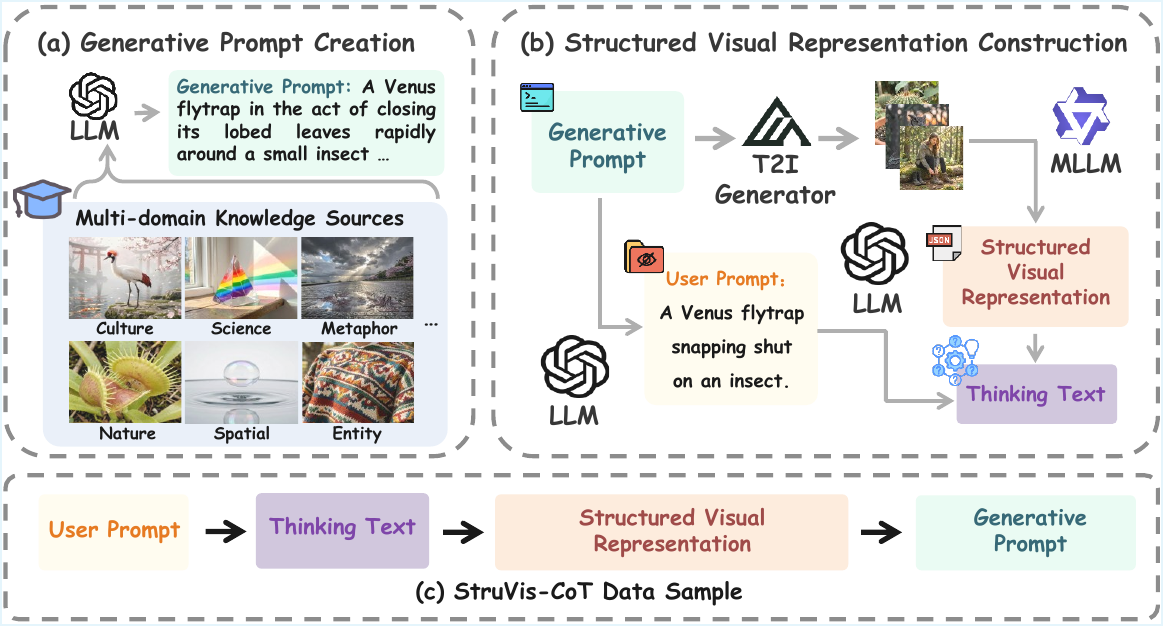}
    \vspace{-12pt}
    \caption{The overall of the data pipeline for collecting our StruVis-CoT data.}
    \vspace{-12pt}
    \label{fig: data}
\end{figure}

\subsection{Data Construction}
To train our proposed \textit{``Thinking with Structured Vision''} model, we construct a specialized dataset, StruVis-CoT, which integrates structured visual representation into the Chain-of-Thought data.
As illustrated in Fig.~\ref{fig: data}, the data construction process involves multiple steps, described as follows:

\noindent \textbf{Generative Prompt Creation.}
We begin by utilizing GPT~\cite{openai2023gpt4} to generate diverse generative prompts $P^T$ across various domains, including culture, nature, science, metaphor, spatial, textual, entity, and story. These prompts are designed to include complex visual scenes, each of which carries intricate constraints such as spatial arrangements, object counts, and attributes. The goal is to capture the richness and diversity of prompts that require reasoning.
    
\noindent \textbf{Image Generation and Structured Visual Representation.} 
We then use the FLUX.2-klein-9B model to generate corresponding images for each textual prompt. FLUX.2-klein-9B is a high-capacity image generator that produces high-quality images conditioned on the user prompts. Once the images are generated, we apply Qwen3-VL-Plus~\cite{bai2025qwen3} to interpret them and extract structured visual representations $S$, including object entities, object relationships, and spatial layouts. These representations are serialized into JSON-formatted text.
    
\noindent \textbf{User Prompt and Thinking Text Generation.}
To abstract the generative prompts into a more generalized form, we use GPT to transform each generative prompt into a less explicit user prompt $P$, which is more ambiguous and concise. Additionally, GPT generates the intermediate Thinking Text $T$ from the inputs of the generated image, which serves as a bridge between the user prompt and the structured visual representation, helping the model reason about the generation process in a more interpretable way.
    
\noindent \textbf{CoT Data Composition.}
Finally, we assemble the StruVis-CoT data $CoT$ into a structured sequence, where each entry consists of the following:
\begin{equation}
CoT = \{ P, T, S, P^T \}.
\end{equation}
The StruVis-CoT is used to train our proposed StruVis, ensuring that it learns how to reason for T2I generation powered by thinking with structured vision.

\subsection{The Proposed StruVis Framework}
In this section, we present \textit{StruVis}, a novel framework designed to enhance reasoning-based T2I generation by introducing structured visual representations. StruVis avoids the limitations of traditional Text-Only and Text-Image Interleaved reasoning frameworks, offering a more efficient and scalable solution. 

\noindent \textbf{Overview of StruVis.}
StruVis operates by representing intermediate reasoning states using text-based structured visual representations instead of relying on intermediate image generation. This method avoids the inefficiencies and constraints posed by repeated image generator calls and the limitations of the image generator. The core idea of our proposed StruVis is to guide the model through reasoning steps that incorporate structured visual representation, ensuring that the generated image respects both the constraints in the prompt and the visual consistency required for high-quality T2I generation.
To implement the “Thinking with Structured Vision” framework for reasoning-based T2I generation, we train StruVis in two stages: \textbf{(i) SFT} and \textbf{(ii) GRPO}.

\noindent \textbf{Training StruVis: SFT.}
In this phase, we fine-tune the MLLM on the StruVis-CoT dataset. The model is trained to adapt to the Chain-of-Thought reasoning format that incorporates structured visual representations. The loss function for this stage is based on minimizing the difference between the model's predicted reasoning processing and the ground truth CoT:
\begin{equation}
\mathcal{L}_{\text{SFT}} = - \sum_{i=1}^{n} \log p_\theta(r_i \mid P , r_{<i}),
\end{equation}
 where $p_\theta$ denotes the policy model parameterized by $\theta$, $P$ is the user prompt, and \( r_i \) is the ground truth reasoning step from the StruVis-CoT data.

\begin{figure}[t!]
    \centering
    \includegraphics[width=\textwidth]{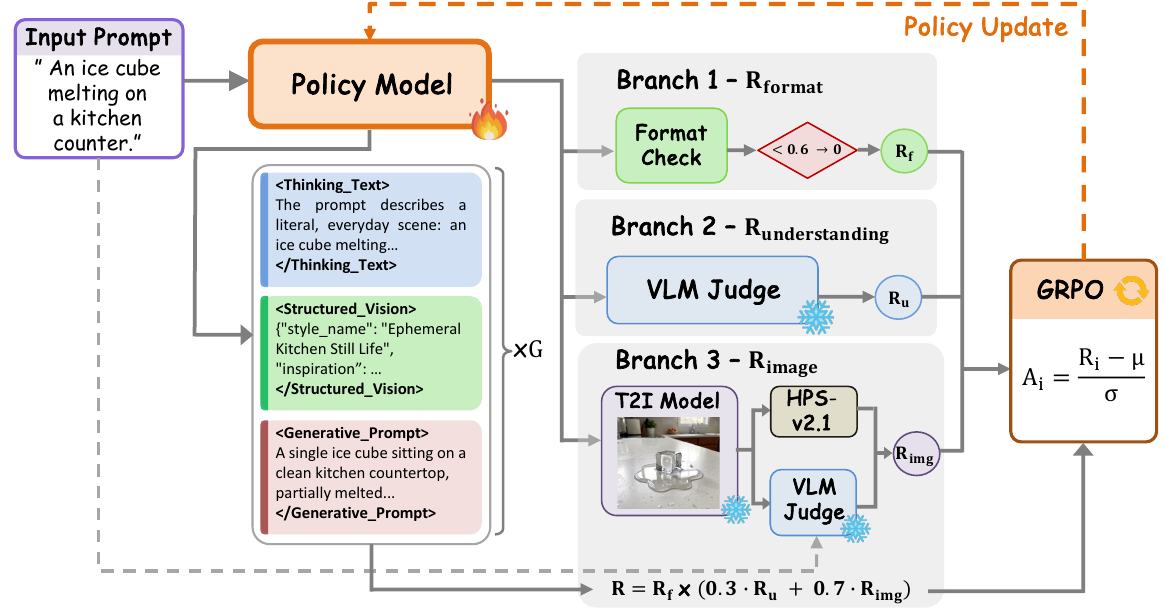}
    \vspace{-12pt}
    \caption{The overall of the GRPO training stage. We design three reward functions to train our StruVis, including format, understanding, and image rewards.}
    \vspace{-12pt}
    \label{fig: grpo}
\end{figure}

\noindent \textbf{Training StruVis: GRPO.}
As shown in Fig.~\ref{fig: grpo}, the second phase of training uses GRPO to further refine the model's reasoning abilities. The reward function is designed to evaluate both the reasoning quality and the image generation quality. We define the rewards as follows:
    
\noindent \textbf{Format Reward} $R_{\mathrm{format}}$.
We assess whether the model output conforms to the required response schema in both presence and well-formedness. Specifically, the reward is computed as a weighted combination of three criteria: (i) Label Completeness $R_{\mathrm{label}}$ , requiring the simultaneous presence of the \texttt{<structure vision>} and \texttt{<final prompt>} tags; (ii) JSON Validity $R_{\mathrm{json}}$, requiring the content enclosed by \texttt{<structure vision>} to be syntactically valid; and (iii) Prompt Validity $R_{\mathrm{prompt}}$, requiring the \texttt{<final prompt>} field to be non-empty.
\begin{equation}
R_{\mathrm{format}}
= 0.4\cdot R_{\mathrm{label}}
+ 0.4\cdot R_{\mathrm{json}}
+ 0.2\cdot R_{\mathrm{prompt}}.
\end{equation}
        
\noindent \textbf{Understanding Reward} \( R_{\mathrm{understanding}} \).
Measures the model's understanding of the original prompt. It is computed as the average of three dimensions ($R_\mathrm{perception}$, $R_\mathrm{completeness}$, $R_\mathrm{faithfulness}$), each scored from 0 to 2:
\begin{equation}
R_{\mathrm{understanding}} = \frac{R_\mathrm{perception} + R_\mathrm{completeness} + R_\mathrm{faithfulness}}{6}.
\end{equation}
\noindent \textbf{Image Reward} $R_{\mathrm{image}}$. We quantify both the perceptual quality of the generated image and its fidelity to the conditioning prompt via a weighted aggregation of two complementary signals: (i) Human Preference Score $R_{\mathrm{HPS}}$, capturing overall visual quality and prompt compliance, and (ii) VLM-based score $R_{\mathrm{VLM}}$, measuring semantic consistency between the image and the prompt.
\begin{equation}
    R_{\mathrm{image}} = 0.6\cdot R_{\mathrm{HPS}} + 0.4\cdot R_{\mathrm{VLM}}.
\end{equation}

\noindent \textbf{Final Reward.} The final reward $R_{\mathrm{final}}$ is computed as:
\begin{equation}
R_{\mathrm{final}} = \mathrm{Gate}(R_{\mathrm{format}}) \cdot  (0.3 \cdot R_{\mathrm{understanding}} + 0.7 \cdot R_{\mathrm{image}}),
\end{equation}

where the gate function ensures that if $R_{\mathrm{format}}$ is less than 0.6, the pipeline terminates early and does not execute downstream steps, including image generation, because the input fails the formatting check.

\subsection{Implementation}

\noindent \textbf{StruVis-CoT Data.}
The data is constructed of 32,599 CoT samples in 8 domains, including culture, nature, science, metaphor, spatial, textual, entity, and story. 

\noindent \textbf{Base Model.}
We utilize two MLLMs, Qwen2.5-VL-7B~\cite{bai2025qwen2} and Qwen3-VL-8B~\cite{bai2025qwen3}, as base models for our StruVis framework. Both base models are trained using the same training strategy and hyperparameters, and both exhibit significant performance improvements in reasoning-based T2I generation.


\noindent \textbf{Reward Model.}
We use the Qwen2.5-VL-72B~\cite{bai2025qwen2} as the reward model for GRPO training. 

\noindent \textbf{Training Details.}
During SFT, we train the base model on the StruVis-CoT dataset with a learning rate of 5e-5. Subsequently, during GRPO, we train the model with a learning rate of 2e-5 and a group size of 8. Cosine learning-rate decay is applied in both stages. All experiments are run on 16 GPUs.

\section{Experiments}

\subsection{Reasoning-based T2I Generation Benchmarks}
We evaluate StruVis on two reasoning-oriented text-to-image (T2I) generation benchmarks: WISE benchmark~\cite{niu2025wise} and T2I-ReasonBench~\cite{sun2025t2i}.
(i) WISE is a world-knowledge-informed benchmark for semantic evaluation of T2I models, featuring 1,000 carefully curated prompts across 25 subdomains spanning cultural common sense, spatio-temporal reasoning, and natural science.
(ii) T2I-ReasonBench is a reasoning-focused benchmark for text-to-image generation, consisting of 800 carefully designed prompts spanning four dimensions: Idiom Interpretation, Textual Image Design, Entity-Reasoning, and Scientific-Reasoning. It adopts a two-stage, prompt-dependent evaluation protocol that measures both reasoning accuracy and image quality via automatically generated question–criterion pairs and multimodal scoring.

\subsection{Quantitative Results}

\begin{table}[t!]
\renewcommand{\tabcolsep}{2pt}
\caption{\textbf{Results on the T2I-ReasonBench with Flux-dev-1}. Accuracy and Quality are reported for each category and overall. Gains are shown in \textcolor{violet!85}{purple}.}
\vspace{-2pt}
\resizebox{\linewidth}{!}{%
\begin{tabular}{
>{\raggedright\arraybackslash}p{1.8cm}| 
>{\centering\arraybackslash}p{1.2cm} 
>{\centering\arraybackslash}p{1.2cm}| 
>{\centering\arraybackslash}p{1.2cm} 
>{\centering\arraybackslash}p{1.2cm}| 
>{\centering\arraybackslash}p{1.2cm} 
>{\centering\arraybackslash}p{1.2cm}| 
>{\centering\arraybackslash}p{1.2cm} 
>{\centering\arraybackslash}p{1.2cm}| 
>{\centering\arraybackslash}p{1.2cm} 
>{\centering\arraybackslash}p{1.2cm}  
}
\toprule
\textbf{Method} & \multicolumn{2}{c|}{\textbf{Idiom}} & \multicolumn{2}{c|}{\textbf{Entity}} & \multicolumn{2}{c|}{\textbf{Scientific}} & \multicolumn{2}{c|}{\textbf{Textual}} & \multicolumn{2}{c}{\textbf{Overall}} \\
 & \textbf{Acc.} & \textbf{Qual.} & \textbf{Acc.} & \textbf{Qual.} & \textbf{Acc.} & \textbf{Qual.} & \textbf{Acc.} & \textbf{Qual.} & \textbf{Acc.} & \textbf{Qual.} \\
\midrule
\multicolumn{11}{c}{\textit{Qwen2.5-VL-7B}} \\
\midrule
W/O. & 46.32 & 90.58 & 59.96 & 94.48 & 58.71 & 88.33 & 64.95 & 78.50 & 57.48 & 87.97 \\
Text-Only & 51.71 & 90.31 & 61.16 & 95.90 & 63.87 & 89.75 & 72.90 & 84.54 & 62.41 & 90.12 \\
Interleaved & 61.46 & 91.79 & 63.55 & 95.98 & 67.48 & 91.08 & 72.90 & 84.54 & 66.35 & 90.85 \\
\rowcolor{gray!10} \textbf{Ours} & \textbf{69.85} & \textbf{93.08} & \textbf{76.30} & \textbf{96.83} & \textbf{72.09} & \textbf{93.29} & \textbf{74.64} & \textbf{84.75} & \textbf{73.22} & \textbf{91.99} \\
\midrule
\rowcolor{gray!20} $\Delta$ & \upx{8.39} & \upx{1.29} & \upx{12.75} & \upx{0.85} & \upx{4.61} & \upx{2.21} & \upx{1.74} & \upx{0.21} & \upx{6.87} & \upx{1.14} \\
\midrule
\multicolumn{11}{c}{\textit{Qwen3-VL-8B}} \\
\midrule
W/O. & 60.66 & 90.77 & 59.25 & 94.13 & 61.77 & 90.42 & 71.35 & 82.71 & 63.26 & 89.51 \\
Text-Only  & 61.43 & 91.08 & 62.03 & 94.48 & 62.40 & 88.25 & 74.03 & 82.96 & 64.97 & 89.19 \\
Interleaved  & 61.92 & 92.31 & 70.56 & 95.04 & 68.46 & 91.75 & 74.92 & 84.71 & 68.97 & 90.95 \\
\rowcolor{gray!10} \textbf{Ours} & \textbf{70.15} & \textbf{92.99} & \textbf{76.74} & \textbf{97.33} & \textbf{72.18} & \textbf{92.12} & \textbf{75.22} & \textbf{84.75} & \textbf{73.57} & \textbf{91.80} \\
\midrule
\rowcolor{gray!20}  $\Delta$ & \upx{8.23} & \upx{0.68} & \upx{6.18} & \upx{2.29} & \upx{3.72} & \upx{0.37} & \upx{0.30} & \upx{0.04} & \upx{4.61} & \upx{0.84} \\
\bottomrule
\end{tabular}%
}
\label{tab:t2ireason_results}
\vspace{-12pt}
\end{table}

\begin{table*}[t!]
\renewcommand{\tabcolsep}{8pt}
\caption{\textbf{The Results on WISE with Flux-dev-1}. We report category-wise scores and the overall score. Gains are shown in \textcolor{violet!85}{purple}.}
\vspace{-4pt}
\resizebox{\linewidth}{!}{
\begin{tabular}{
>{\raggedright\arraybackslash}p{1.8cm}|  
>{\centering\arraybackslash}p{1.2cm}     
>{\centering\arraybackslash}p{1.2cm}     
>{\centering\arraybackslash}p{1.2cm}     
>{\centering\arraybackslash}p{1.2cm}     
>{\centering\arraybackslash}p{1.2cm}     
>{\centering\arraybackslash}p{1.2cm}     
>{\centering\arraybackslash}p{1.2cm}     
}
\toprule
\textbf{Method} & \textbf{Cultural} & \textbf{Time} & \textbf{Space} & \textbf{Biology} & \textbf{Physics} & \textbf{Chemistry} & \textbf{Overall} \\
\midrule

\multicolumn{8}{c}{\textit{Qwen2.5-VL-7B}} \\
\midrule
W/O.         & 0.62 & 0.49 & 0.61 & 0.50 & 0.54 & 0.41 & 0.56 \\
Text-Only    & 0.66 & 0.51 & 0.62 & 0.52 & 0.51 & 0.47 & 0.58 \\
Interleaved  & 0.61 & 0.50 & 0.62 & 0.50 & 0.52 & 0.42 & 0.55 \\
\rowcolor{gray!10} Ours & \textbf{0.75} & \textbf{0.59} & \textbf{0.63} & \textbf{0.58} & \textbf{0.58} & \textbf{0.51} & \textbf{0.65} \\
\rowcolor{gray!20} $\Delta$ & \upx{0.14} & \upx{0.09} & \upx{0.01} & \upx{0.08} & \upx{0.06} & \upx{0.09} & \upx{0.10} \\

\midrule
\multicolumn{8}{c}{\textit{Qwen3-VL-8B}} \\
\midrule
W/O.         & 0.61 & 0.45 & 0.61 & 0.44 & 0.52 & 0.41 & 0.53 \\
Text-Only    & 0.62 & 0.49 & 0.61 & 0.50 & 0.56 & 0.44 & 0.56 \\
Interleaved  & 0.62 & 0.55 & 0.61 & 0.53 & 0.51 & 0.47 & 0.57 \\
\rowcolor{gray!10} Ours & \textbf{0.74} & \textbf{0.59} & \textbf{0.62} & \textbf{0.59} & \textbf{0.56} & \textbf{0.51} & \textbf{0.65} \\
\rowcolor{gray!20} $\Delta$ & \upx{0.12} & \upx{0.04} & \upx{0.01} & \upx{0.06} & \upx{0.05} & \upx{0.04} & \upx{0.08} \\

\bottomrule
\end{tabular}
}
\label{tab:wise_ablation}
\vspace{-12pt}
\end{table*}

\noindent \textbf{Results on T2I-ReasonBench.}
As shown in Table~\ref{tab:t2ireason_results}, we evaluate StruVis on T2I-ReasonBench under two base models, Qwen2.5-VL-7B and Qwen3-VL-8B, and report both accuracy (acc.) and perceptual quality (qual.) across four categories, \eg, Idiom, Entity, Scientific, and Textual. 
Overall, StruVis consistently achieves the best performance, demonstrating that thinking with structured vision provides an effective improvement for T2I reasoning.

On Qwen2.5-VL-7B, StruVis improves the overall accuracy from 66.35 to \textbf{73.22} (\textbf{+6.87}), while also increasing overall quality from 90.85 to 91.99 (+1.14). Notably, the gains are most pronounced on the Entity category, where StruVis boosts accuracy from 63.55 to \textbf{76.30} (\textbf{+12.75}), indicating substantially better preservation of object inventories, attribute bindings, and relation constraints, precisely the failure modes commonly observed in text-only reasoning without visual interaction. StruVis also yields significant improvements on Idiom with a performance gain of 8.39\%, suggesting that the proposed structured visual intermediate state benefits both compositional reasoning and semantically grounded constraint execution. Similar trends hold for the stronger Qwen3-VL-8B backbone, where StruVis further raises overall accuracy from 68.97 to 73.57 with an improvement of \textbf{4.61\%}. It also achieves the highest overall quality score.

Compared to Text-Only Reasoning, which offers limited gains due to the lack of explicit visual context, and Text–Image Interleaved Reasoning, which is constrained by repeated T2I generator calls and the representational bottleneck of intermediate image generation, StruVis attains superior constraint satisfaction without relying on costly or unreliable intermediate images. These results validate the effectiveness of StruVis by introducing a text-based structured visual representation as the intermediate reasoning state. 

\noindent \textbf{Results on WISE.}
As shown in Table~\ref{tab:wise_ablation}, StruVis consistently yields the best performance on WISE across both MLLMs, highlighting the effectiveness of thinking with structured vision for T2I reasoning. With Qwen2.5-VL-7B, StruVis improves the overall score by \textbf{+0.10\%} (from 0.55/0.58 to 0.65), with the largest gains on Cultural (\textbf{+0.14\%}) and Time/Chemistry (both \textbf{+0.09\%}), indicating substantially stronger grounding in cultural and temporal instructions, as well as chemistry-related knowledge. Similarly, on Qwen3-VL-8B, StruVis increases the overall score by \textbf{+0.08\%} (0.57 to 0.65), driven primarily by Cultural (\textbf{+0.12\%}) and Biology (\textbf{+0.06\%}) improvements. These consistent and pronounced gains suggest that replacing intermediate image generations with text-based structured visual states provides a more reliable and efficient reasoning interface, enabling better knowledge grounding and instruction faithfulness.

\subsection{Qualitative Results}

\begin{figure}[t!]
    \centering
    \includegraphics[width=\textwidth]{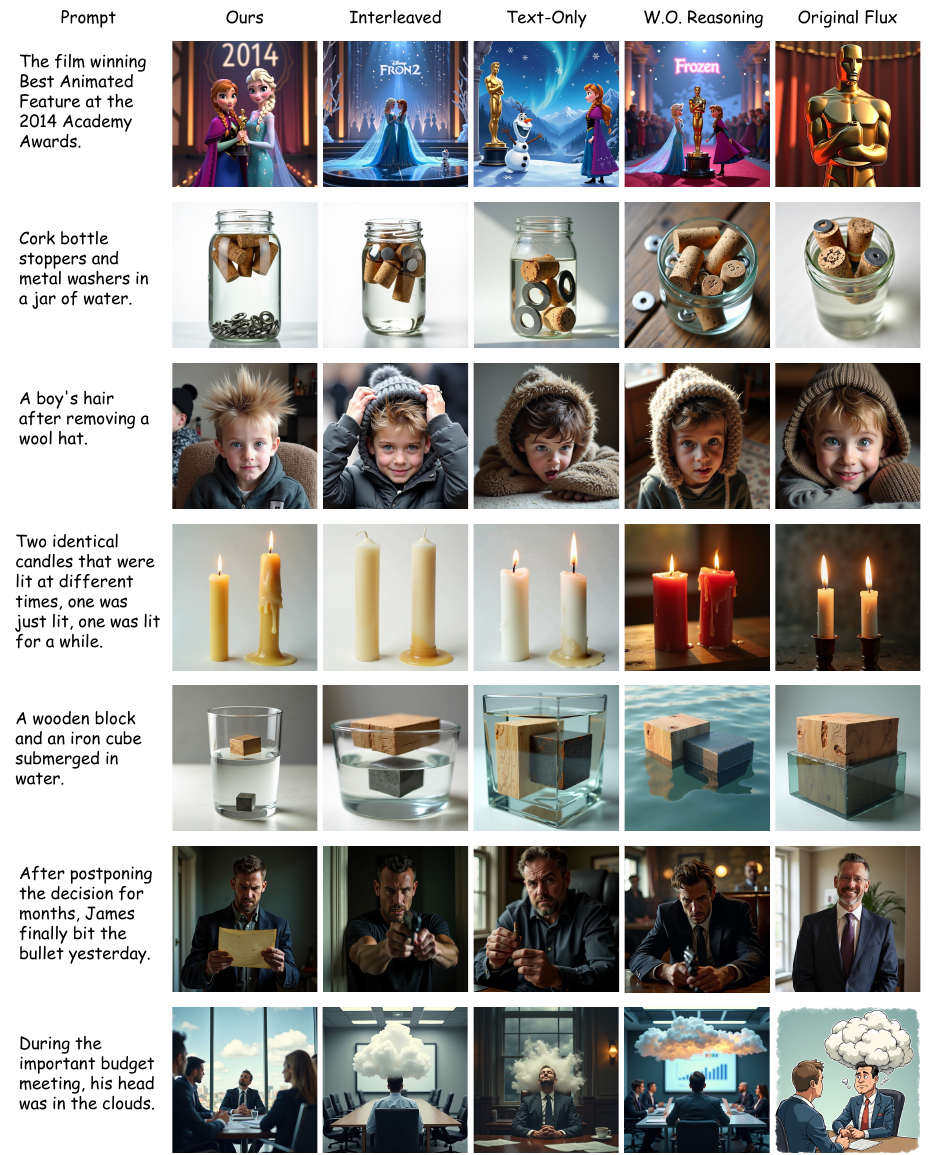}
    \vspace{-12pt}
    \caption{The visual comparison of our proposed StruVis and the baselines. We show the final generated results on the T2I-ReasonBench.}
    \vspace{-20pt}
    \label{fig: t2ireason}
\end{figure}

\begin{figure}[t!]
    \centering
    \includegraphics[width=\textwidth]{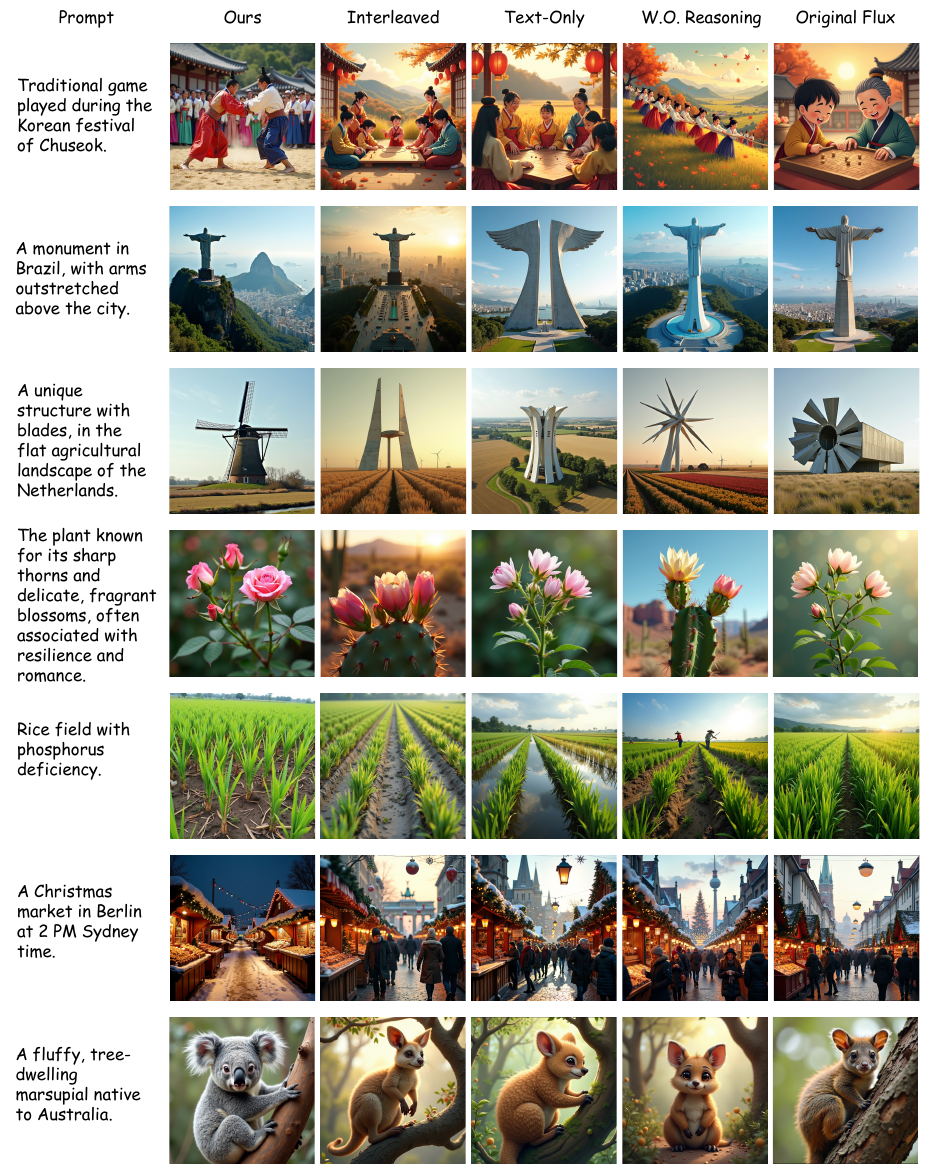}
    \vspace{-12pt}
    \caption{The visual comparison of our proposed StruVis and the baselines. We show the final generated results on the WISE Benchmark.}
    \vspace{-28pt}
    \label{fig: wise}
\end{figure}

Fig.~\ref{fig: t2ireason} presents a qualitative comparison of seven challenging prompts from T2I-ReasonBench. Overall, StruVis shows significant performance gains compared with strong baselines, consistently producing images that better satisfy multi-object constraints, \eg, correct object counts, attribute bindings, and spatial layouts, while preserving visual coherence. By introducing text-based structured visual representations as intermediate reasoning states, StruVis enhances reasoning-based T2I generation without relying on costly and failure-prone intermediate image generation, yielding both improved effectiveness and practicality.

For example, in the fourth case \textit{``Two identical candles that were lit at different times''}, StruVis correctly captures the relative temporal state by rendering one candle noticeably more burned with plausible wax dripping, while keeping the two candles visually consistent as ``identical'' objects. In contrast, Text-Only Reasoning often collapses the temporal constraint and produces candles of similar height or ambiguous burn traces, and interleaved reasoning may still drift due to imperfect intermediate generations. In the fifth case \textit{``A wooden block and an iron cube submerged in water''}, StruVis demonstrates stronger commonsense grounding by placing the wooden block floating near the water surface and the iron cube sinking, with a coherent waterline and stable spatial configuration. Baselines frequently violate buoyancy-related constraints or produce unstable compositions, indicating that they struggle to reliably translate physical priors into globally consistent layouts. In the seventh case \textit{``During the important budget meeting, his head was in the clouds.''}, StruVis better resolves figurative language into a visually grounded depiction that simultaneously preserves the meeting context and conveys the metaphor in a controlled manner, whereas competing methods tend to either literalize incorrectly, ignore the idiom, or introduce visually inconsistent elements that break scene semantics.

\begin{wraptable}{r}{0.6\textwidth}
\renewcommand{\tabcolsep}{8pt}
\vspace{-24pt}
\caption{The ablation results on T2I-ReasonBench to demonstrate the effectiveness of the proposed three reward functions.}
\resizebox{\linewidth}{!}{
\begin{tabular}{c|ccc|cc}
\toprule
\multicolumn{1}{c|}{SFT} &
\multicolumn{3}{c|}{GRPO} &
\multicolumn{2}{c}{T2I-ReasonBench} \\
\cmidrule{2-4}\cmidrule{5-6}
& $R_{\mathrm{format}}$ & $R_{\mathrm{understanding}}$ & $R_{\mathrm{image}}$ & Acc. & Qual. \\
\midrule

\multicolumn{6}{c}{\textit{Qwen2.5-VL-7B}} \\
\midrule
\cmark &  \xmark & \xmark  & \xmark  & 62.43 & 89.41 \\
\cmark & \cmark & \xmark  & \xmark  & 69.10 & 90.56 \\
\cmark & \cmark & \cmark & \xmark  & 71.33 & 90.78 \\
\cmark & \cmark & \cmark & \cmark & 73.22 & 91.99 \\
\midrule

\multicolumn{6}{c}{\textit{Qwen3-VL-8B}} \\
\midrule
\cmark &  \xmark & \xmark  & \xmark  & 66.62 & 89.40 \\
\cmark & \cmark & \xmark  & \xmark  & 68.01 & 90.57 \\
\cmark & \cmark & \cmark & \xmark  & 71.35 & 91.53 \\
\cmark & \cmark & \cmark & \cmark & 73.57 & 91.80 \\
\bottomrule
\end{tabular}}
\vspace{-16pt}
\label{tab: ab_reward}
\end{wraptable}

Additionally, we show the visual comparison on the WISE benchmark in Fig.~\ref{fig: wise}. 
For instance, in the prompt \textit{``A monument in Brazil, with arms outstretched above the city.''}, StruVis correctly renders an iconic statue-on-mountain composition consistent with the intended landmark, together with a believable cityscape background. In contrast, competing methods tend to generate generic statues or structurally incorrect monuments, indicating weaker grounding to the specific concept implied by the prompt. 
Furthermore, for \textit{``A Christmas market in Berlin at 2 PM Sydney time.''}, StruVis better aligns the global illumination and scene atmosphere with the implied time conversion (daytime in Berlin rather than a nighttime market appearance), whereas baselines often default to stereotypical nighttime Christmas-market imagery, ignoring the temporal constraint.

Overall, all qualitative results highlight that StruVis more faithfully follows complex instructions that require state reasoning, physical plausibility, and non-literal language grounding, \emph{demonstrating strong qualitative effectiveness for reasoning-based T2I generation across diverse constraint types.}

\section{Ablation Study}

\subsection{Ablation of Reward Functions}

As shown in Table~\ref{tab: ab_reward}, we report the ablation study on T2I-ReasonBench, validating the effectiveness of the three reward functions used in our GRPO stage, \ie, $R_{\text{format}}$, $R_{\text{understanding}}$, and $R_{\text{image}}$. We conduct controlled comparisons by progressively enabling these rewards on top of the same SFT initialization, and evaluate both reasoning accuracy and overall generation quality across two base models (Qwen2.5-VL-7B and Qwen3-VL-8B).
Overall, we observe a clear and consistent trend: that adding each reward yields monotonic improvements, and using all three rewards achieves the best performance. The combination of the three reward functions provides the most effective alignment objective for GRPO, leading to the strongest and most consistent gains across base models.

\subsection{MLLM \textit{vs.} LLM}

\begin{wrapfigure}{t}{0.6\textwidth} 
  \centering
  \includegraphics[width=\linewidth]{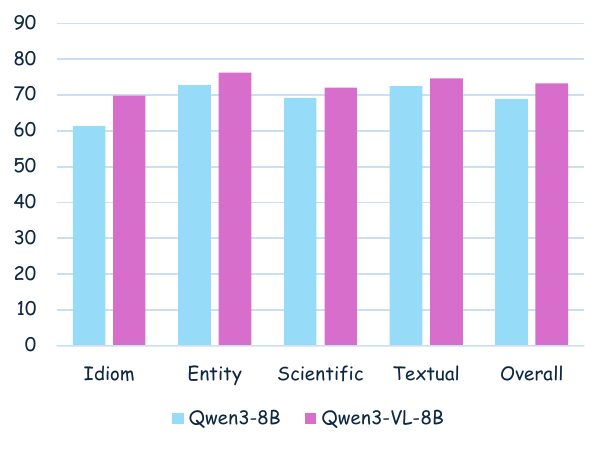}
    \vspace{-20pt}
    \caption{We report the performance comparison on T2I-ReasonBench between the LLM-based StruVis and the MLLM-based StruVis.
    }
    \vspace{10pt}
    \label{fig: LLMvsMLLM}
\end{wrapfigure}

Finally, we conduct an interesting analysis motivated by the fact that our CoT data is purely text-formatted, which in principle allows StruVis to be trained on either an LLM or an MLLM. This raises a natural question: \textit{how does an LLM-based StruVis (\eg, Qwen3-8B~\cite{yang2025qwen3}) compare with an MLLM-based StruVis (\eg, Qwen3-VL-8B) under the same structured-vision reasoning?} To answer this, we build a controlled ablation and report the results in Fig.~\ref{fig: LLMvsMLLM}.
The results show that StruVis trained on Qwen3-VL-8B consistently and substantially outperforms its Qwen3-8B counterpart on T2I-ReasonBench.
The results suggest that learning \textbf{\textit{“structured vision”}} for reasoning is not an exclusively symbolic process. \emph{However, effective intermediate visual-state modeling benefits from a certain degree of built-in visual knowledge, which helps bridge structured textual representations and their realizable visual counterparts during T2I generation.}

\section{Conclusion}
In this paper, we presented \textbf{StruVis}, a new reasoning framework for reasoning-based text-to-image generation that enables \textit{thinking with structured vision}. Unlike existing Text-Only Reasoning that lacks visual interaction, or Text-Image Interleaved Reasoning that relies on repeated image generation, StruVis introduces the text-based structured visual representation as an intermediate reasoning state. 
To this end, we constructed \textbf{StruVis-CoT}, a \textit{\textbf{``Text-Structured Vision''}} interleaved chain-of-thought data, and trained MLLMs with SFT and GRPO to align them with the proposed reasoning method. Extensive experiments on reasoning-based T2I benchmarks demonstrate consistent gains, including \textbf{+4.61\%} on T2I-ReasonBench and \textbf{+4\%} on WISE, validating both the effectiveness and practicality of StruVis as a model-agnostic enhancement.


%
%
\bibliographystyle{splncs04}
\bibliography{main}

@article{shao2024deepseekmath,
  title={Deepseekmath: Pushing the limits of mathematical reasoning in open language models},
  author={Shao, Zhihong and Wang, Peiyi and Zhu, Qihao and Xu, Runxin and Song, Junxiao and Bi, Xiao and Zhang, Haowei and Zhang, Mingchuan and Li, YK and Wu, Yang and others},
  journal={arXiv preprint arXiv:2402.03300},
  year={2024}
}

@article{guo2025deepseek,
  title={Deepseek-r1: Incentivizing reasoning capability in llms via reinforcement learning},
  author={Guo, Daya and Yang, Dejian and Zhang, Haowei and Song, Junxiao and Wang, Peiyi and Zhu, Qihao and Xu, Runxin and Zhang, Ruoyu and Ma, Shirong and Bi, Xiao and others},
  journal={arXiv preprint arXiv:2501.12948},
  year={2025}
}

@article{yang2025qwen25,
  title={Qwen2.5 Technical Report},
  author={Yang, An and Yang, Baosong and Zhang, Beichen and Hui, Binyuan and Zheng, Bo and Yu, Bowen and Li, Chengyuan and Liu, Dayiheng and Huang, Fei and Wei, Haoran and others},
  journal={arXiv preprint arXiv:2412.15115},
  year={2025}
}

@article{yang2025qwen3,
  title={Qwen3 technical report},
  author={Yang, An and Li, Anfeng and Yang, Baosong and Zhang, Beichen and Hui, Binyuan and Zheng, Bo and Yu, Bowen and Gao, Chang and Huang, Chengen and Lv, Chenxu and others},
  journal={arXiv preprint arXiv:2505.09388},
  year={2025}
}

@article{he2025skywork,
  title={Skywork open reasoner 1 technical report},
  author={He, Jujie and Liu, Jiacai and Liu, Chris Yuhao and Yan, Rui and Wang, Chaojie and Cheng, Peng and Zhang, Xiaoyu and Zhang, Fuxiang and Xu, Jiacheng and Shen, Wei and others},
  journal={arXiv preprint arXiv:2505.22312},
  year={2025}
}

@article{xie2025logic,
  title={Logic-rl: Unleashing llm reasoning with rule-based reinforcement learning},
  author={Xie, Tian and Gao, Zitian and Ren, Qingnan and Luo, Haoming and Hong, Yuqian and Dai, Bryan and Zhou, Joey and Qiu, Kai and Wu, Zhirong and Luo, Chong},
  journal={arXiv preprint arXiv:2502.14768},
  year={2025}
}

@article{yu2025dapo,
  title={Dapo: An open-source llm reinforcement learning system at scale},
  author={Yu, Qiying and Zhang, Zheng and Zhu, Ruofei and Yuan, Yufeng and Zuo, Xiaochen and Yue, Yu and Dai, Weinan and Fan, Tiantian and Liu, Gaohong and Liu, Lingjun and others},
  journal={arXiv preprint arXiv:2503.14476},
  year={2025}
}

@article{liu2025drgrpo,
  title={Understanding r1-zero-like training: A critical perspective},
  author={Liu, Zichen and Chen, Changyu and Li, Wenjun and Qi, Penghui and Pang, Tianyu and Du, Chao and Lee, Wee Sun and Lin, Min},
  journal={arXiv preprint arXiv:2503.20783},
  year={2025}
}

@article{zhang2025critiquegrpo,
  title={Critique-grpo: Advancing llm reasoning with natural language and numerical feedback},
  author={Zhang, Xiaoying and Sun, Hao and Zhang, Yipeng and Feng, Kaituo and Lu, Chaochao and Yang, Chao and Meng, Helen},
  journal={arXiv preprint arXiv:2506.03106},
  year={2025}
}

@article{xue2025dancegrpo,
  title={Dancegrpo: Unleashing grpo on visual generation},
  author={Xue, Zeyue and Wu, Jie and Gao, Yu and Kong, Fangyuan and Zhu, Lingting and Chen, Mengzhao and Liu, Zhiheng and Liu, Wei and Guo, Qiushan and Huang, Weilin and others},
  journal={arXiv preprint arXiv:2505.07818},
  year={2025}
}

@article{zhang2025thyme,
  title={Thyme: Think beyond images},
  author={Zhang, Yi-Fan and Lu, Xingyu and Yin, Shukang and Fu, Chaoyou and Chen, Wei and Hu, Xiao and Wen, Bin and Jiang, Kaiyu and Liu, Changyi and Zhang, Tianke and others},
  journal={arXiv preprint arXiv:2508.11630},
  year={2025}
}

@article{liu2025seg,
  title={Seg-zero: Reasoning-chain guided segmentation via cognitive reinforcement},
  author={Liu, Yuqi and Peng, Bohao and Zhong, Zhisheng and Yue, Zihao and Lu, Fanbin and Yu, Bei and Jia, Jiaya},
  journal={arXiv preprint arXiv:2503.06520},
  year={2025}
}

@article{fan2025grit,
  title={Grit: Teaching mllms to think with images},
  author={Fan, Yue and He, Xuehai and Yang, Diji and Zheng, Kaizhi and Kuo, Ching-Chen and Zheng, Yuting and Narayanaraju, Sravana Jyothi and Guan, Xinze and Wang, Xin Eric},
  journal={arXiv preprint arXiv:2505.15879},
  year={2025}
}

@article{ni2025point,
  title={Point-rft: Improving multimodal reasoning with visually grounded reinforcement finetuning},
  author={Ni, Minheng and Yang, Zhengyuan and Li, Linjie and Lin, Chung-Ching and Lin, Kevin and Zuo, Wangmeng and Wang, Lijuan},
  journal={arXiv preprint arXiv:2505.19702},
  year={2025}
}

@article{su2025openthinkimg,
  title={Openthinkimg: Learning to think with images via visual tool reinforcement learning},
  author={Su, Zhaochen and Li, Linjie and Song, Mingyang and Hao, Yunzhuo and Yang, Zhengyuan and Zhang, Jun and Chen, Guanjie and Gu, Jiawei and Li, Juntao and Qu, Xiaoye and others},
  journal={arXiv preprint arXiv:2505.08617},
  year={2025}
}

@article{zheng2025deepeyes,
  title={Deepeyes: Incentivizing" thinking with images" via reinforcement learning},
  author={Zheng, Ziwei and Yang, Michael and Hong, Jack and Zhao, Chenxiao and Xu, Guohai and Yang, Le and Shen, Chao and Yu, Xing},
  journal={arXiv preprint arXiv:2505.14362},
  year={2025}
}

@article{wu2025vtool,
  title={Vtool-r1: Vlms learn to think with images via reinforcement learning on multimodal tool use},
  author={Wu, Mingyuan and Yang, Jingcheng and Jiang, Jize and Li, Meitang and Yan, Kaizhuo and Yu, Hanchao and Zhang, Minjia and Zhai, Chengxiang and Nahrstedt, Klara},
  journal={arXiv preprint arXiv:2505.19255},
  year={2025}
}

@article{wang2025pixel,
  title={Pixel reasoner: Incentivizing pixel-space reasoning with curiosity-driven reinforcement learning},
  author={Wang, Haozhe and Su, Alex and Ren, Weiming and Lin, Fangzhen and Chen, Wenhu},
  journal={arXiv preprint arXiv:2505.15966},
  year={2025}
}

@article{bai2025univg,
  title={Univg-r1: Reasoning guided universal visual grounding with reinforcement learning},
  author={Bai, Sule and Li, Mingxing and Liu, Yong and Tang, Jing and Zhang, Haoji and Sun, Lei and Chu, Xiangxiang and Tang, Yansong},
  journal={arXiv preprint arXiv:2505.14231},
  year={2025}
}

@article{huang2025visualtoolagent,
  title={Visualtoolagent (vista): A reinforcement learning framework for visual tool selection},
  author={Huang, Zeyi and Ji, Yuyang and Rajan, Anirudh Sundara and Cai, Zefan and Xiao, Wen and Wang, Haohan and Hu, Junjie and Lee, Yong Jae},
  journal={arXiv preprint arXiv:2505.20289},
  year={2025}
}

@article{zhu2025active,
  title={Active-o3: Empowering multimodal large language models with active perception via grpo},
  author={Zhu, Muzhi and Zhong, Hao and Zhao, Canyu and Du, Zongze and Huang, Zheng and Liu, Mingyu and Chen, Hao and Zou, Cheng and Chen, Jingdong and Yang, Ming and others},
  journal={arXiv preprint arXiv:2505.21457},
  year={2025}
}

@article{yang2025visionthink,
  title={Visionthink: Smart and efficient vision language model via reinforcement learning},
  author={Yang, Senqiao and Li, Junyi and Lai, Xin and Yu, Bei and Zhao, Hengshuang and Jia, Jiaya},
  journal={arXiv preprint arXiv:2507.13348},
  year={2025}
}

@article{sun2025t2i,
  title={T2i-reasonbench: Benchmarking reasoning-informed text-to-image generation},
  author={Sun, Kaiyue and Fang, Rongyao and Duan, Chengqi and Liu, Xian and Liu, Xihui},
  journal={arXiv preprint arXiv:2508.17472},
  year={2025}
}

@article{niu2025wise,
  title={Wise: A world knowledge-informed semantic evaluation for text-to-image generation},
  author={Niu, Yuwei and Ning, Munan and Zheng, Mengren and Jin, Weiyang and Lin, Bin and Jin, Peng and Liao, Jiaqi and Feng, Chaoran and Ning, Kunpeng and Zhu, Bin and others},
  journal={arXiv preprint arXiv:2503.07265},
  year={2025}
}

@article{ye2025visual,
  title={Visual-Aware CoT: Achieving High-Fidelity Visual Consistency in Unified Models},
  author={Ye, Zixuan and Liu, Quande and Wei, Cong and Zhang, Yuanxing and Wang, Xintao and Wan, Pengfei and Gai, Kun and Luo, Wenhan},
  journal={arXiv preprint arXiv:2512.19686},
  year={2025}
}

@article{qin2025uni,
  title={Uni-cot: Towards unified chain-of-thought reasoning across text and vision},
  author={Qin, Luozheng and Gong, Jia and Sun, Yuqing and Li, Tianjiao and Yang, Mengping and Yang, Xiaomeng and Qu, Chao and Tan, Zhiyu and Li, Hao},
  journal={arXiv preprint arXiv:2508.05606},
  year={2025}
}

@article{lyu2025understanding,
  title={Understanding-in-Generation: Reinforcing Generative Capability of Unified Model via Infusing Understanding into Generation},
  author={Lyu, Yuanhuiyi and Wong, Chi Kit and Liao, Chenfei and Jiang, Lutao and Zheng, Xu and Lu, Zexin and Zhang, Linfeng and Hu, Xuming},
  journal={arXiv preprint arXiv:2509.18639},
  year={2025}
}

@inproceedings{gu2025thinkmorph,
  title={Thinkmorph: Emergent properties in multimodal interleaved chain-of-thought reasoning},
  author={Gu, Jiawei and Hao, Yunzhuo and Wang, Huichen Will and Li, Linjie and Shieh, Michael Qizhe and Choi, Yejin and Krishna, Ranjay and Cheng, Yu},
  booktitle={The Fourteenth International Conference on Learning Representations},
  year={2025}
}

@article{jiang2025t2i,
  title={T2i-r1: Reinforcing image generation with collaborative semantic-level and token-level cot},
  author={Jiang, Dongzhi and Guo, Ziyu and Zhang, Renrui and Zong, Zhuofan and Li, Hao and Zhuo, Le and Yan, Shilin and Heng, Pheng-Ann and Li, Hongsheng},
  journal={arXiv preprint arXiv:2505.00703},
  year={2025}
}

@article{zhang2025reasongen,
  title={Reasongen-r1: Cot for autoregressive image generation models through sft and rl},
  author={Zhang, Yu and Li, Yunqi and Yang, Yifan and Wang, Rui and Yang, Yuqing and Qi, Dai and Bao, Jianmin and Chen, Dongdong and Luo, Chong and Qiu, Lili},
  journal={arXiv preprint arXiv:2505.24875},
  year={2025}
}

@article{wang2025pref,
  title={Pref-grpo: Pairwise preference reward-based grpo for stable text-to-image reinforcement learning},
  author={Wang, Yibin and Li, Zhimin and Zang, Yuhang and Zhou, Yujie and Bu, Jiazi and Wang, Chunyu and Lu, Qinglin and Jin, Cheng and Wang, Jiaqi},
  journal={arXiv preprint arXiv:2508.20751},
  year={2025}
}

@article{zhang2025layercraft,
  title={Layercraft: Enhancing text-to-image generation with cot reasoning and layered object integration},
  author={Zhang, Yuyao and Li, Jinghao and Tai, Yu-Wing},
  journal={arXiv preprint arXiv:2504.00010},
  year={2025}
}

@article{huang2025interleaving,
  title={Interleaving reasoning for better text-to-image generation},
  author={Huang, Wenxuan and Chen, Shuang and Xie, Zheyong and Cao, Shaosheng and Tang, Shixiang and Shen, Yufan and Yin, Qingyu and Hu, Wenbo and Wang, Xiaoman and Tang, Yuntian and others},
  journal={arXiv preprint arXiv:2509.06945},
  year={2025}
}

@inproceedings{liao2025imagegen,
  title={Imagegen-cot: Enhancing text-to-image in-context learning with chain-of-thought reasoning},
  author={Liao, Jiaqi and Yang, Zhengyuan and Li, Linjie and Li, Dianqi and Lin, Kevin and Cheng, Yu and Wang, Lijuan},
  booktitle={Proceedings of the IEEE/CVF International Conference on Computer Vision},
  pages={17214--17223},
  year={2025}
}

@article{jiang2025draco,
  title={DraCo: Draft as CoT for Text-to-Image Preview and Rare Concept Generation},
  author={Jiang, Dongzhi and Zhang, Renrui and Li, Haodong and Zong, Zhuofan and Guo, Ziyu and He, Jun and Guo, Claire and Ye, Junyan and Fang, Rongyao and Li, Weijia and others},
  journal={arXiv preprint arXiv:2512.05112},
  year={2025}
}

@article{liu2025cot,
  title={CoT-lized Diffusion: Let's Reinforce T2I Generation Step-by-step},
  author={Liu, Zheyuan and Ning, Munan and Zhang, Qihui and Yang, Shuo and Wang, Zhongrui and Yang, Yiwei and Xu, Xianzhe and Song, Yibing and Chen, Weihua and Wang, Fan and others},
  journal={arXiv preprint arXiv:2507.04451},
  year={2025}
}

@article{guo2025can,
  title={Can We Generate Images with CoT? Let's Verify and Reinforce Image Generation Step by Step},
  author={Guo, Ziyu and Zhang, Renrui and Tong, Chengzhuo and Zhao, Zhizheng and Huang, Rui and Zhang, Haoquan and Zhang, Manyuan and Liu, Jiaming and Zhang, Shanghang and Gao, Peng and others},
  journal={arXiv preprint arXiv:2501.13926},
  year={2025}
}

@article{lin2025decot,
  title={DeCoT: Decomposing Complex Instructions for Enhanced Text-to-Image Generation with Large Language Models},
  author={Lin, Xiaochuan and Chen, Xiangyong and Li, Xuan and Su, Yichen},
  journal={arXiv preprint arXiv:2508.12396},
  year={2025}
}

@article{gu2025improving,
  title={Improving Chain-of-Thought Efficiency for Autoregressive Image Generation},
  author={Gu, Zeqi and Georgopoulos, Markos and Dai, Xiaoliang and Ghazvininejad, Marjan and Wang, Chu and Juefei-Xu, Felix and Li, Kunpeng and Shi, Yujun and He, Zecheng and He, Zijian and others},
  journal={arXiv preprint arXiv:2510.05593},
  year={2025}
}

@article{li2025visual,
  title={Visual-CoG: Stage-Aware Reinforcement Learning with Chain of Guidance for Text-to-Image Generation},
  author={Li, Yaqi and Chen, Peng and Han, Mingyang and Bu, Pi and Shi, Haoxiang and Zhao, Runzhou and Yao, Yang and Zhang, Xuan and Song, Jun and Zheng, Bo},
  journal={arXiv preprint arXiv:2508.18032},
  year={2025}
}

@article{zhou2025reinforced,
  title={Reinforced mllm: A survey on rl-based reasoning in multimodal large language models},
  author={Zhou, Guanghao and Qiu, Panjia and Chen, Cen and Wang, Jie and Yang, Zheming and Xu, Jian and Qiu, Minghui},
  journal={arXiv preprint arXiv:2504.21277},
  year={2025}
}

@article{liu2025flow,
  title={Flow-grpo: Training flow matching models via online rl},
  author={Liu, Jie and Liu, Gongye and Liang, Jiajun and Li, Yangguang and Liu, Jiaheng and Wang, Xintao and Wan, Pengfei and Zhang, Di and Ouyang, Wanli},
  journal={arXiv preprint arXiv:2505.05470},
  year={2025}
}

@misc{flux,
    title = {Black Forest Labs; Frontier AI Lab},
    url = {https://blackforestlabs.ai/},
    author = {BlackForest},
    year = {2024}
}

@misc{openai2023gpt4,
      title={GPT-4 Technical Report}, 
      author={OpenAI},
      year={2023},
      eprint={2303.08774},
      archivePrefix={arXiv},
      primaryClass={cs.CL}
}

@article{wu2025qwen,
  title={Qwen-image technical report},
  author={Wu, Chenfei and Li, Jiahao and Zhou, Jingren and Lin, Junyang and Gao, Kaiyuan and Yan, Kun and Yin, Sheng-ming and Bai, Shuai and Xu, Xiao and Chen, Yilei and others},
  journal={arXiv preprint arXiv:2508.02324},
  year={2025}
}

@article{cao2025hunyuanimage,
  title={Hunyuanimage 3.0 technical report},
  author={Cao, Siyu and Chen, Hangting and Chen, Peng and Cheng, Yiji and Cui, Yutao and Deng, Xinchi and Dong, Ying and Gong, Kipper and Gu, Tianpeng and Gu, Xiusen and others},
  journal={arXiv preprint arXiv:2509.23951},
  year={2025}
}

@article{deng2025emerging,
  title={Emerging properties in unified multimodal pretraining},
  author={Deng, Chaorui and Zhu, Deyao and Li, Kunchang and Gou, Chenhui and Li, Feng and Wang, Zeyu and Zhong, Shu and Yu, Weihao and Nie, Xiaonan and Song, Ziang and others},
  journal={arXiv preprint arXiv:2505.14683},
  year={2025}
}

@article{bai2025qwen3,
  title={Qwen3-vl technical report},
  author={Bai, Shuai and Cai, Yuxuan and Chen, Ruizhe and Chen, Keqin and Chen, Xionghui and Cheng, Zesen and Deng, Lianghao and Ding, Wei and Gao, Chang and Ge, Chunjiang and others},
  journal={arXiv preprint arXiv:2511.21631},
  year={2025}
}

@article{bai2025qwen2,
  title={Qwen2. 5-VL Technical Report},
  author={Bai, Shuai and Chen, Keqin and Liu, Xuejing and Wang, Jialin and Ge, Wenbin and Song, Sibo and Dang, Kai and Wang, Peng and Wang, Shijie and Tang, Jun and others},
  journal={arXiv e-prints},
  pages={arXiv--2502},
  year={2025}
}

@article{zhang2025latent,
  title={Latent Sketchpad: Sketching Visual Thoughts to Elicit Multimodal Reasoning in MLLMs},
  author={Zhang, Huanyu and Wu, Wenshan and Li, Chengzu and Shang, Ning and Xia, Yan and Huang, Yangyu and Zhang, Yifan and Dong, Li and Zhang, Zhang and Wang, Liang and others},
  journal={arXiv preprint arXiv:2510.24514},
  year={2025}
}

@article{zhang2025scaling,
  title={Scaling and Beyond: Advancing Spatial Reasoning in MLLMs Requires New Recipes},
  author={Zhang, Huanyu and Li, Chengzu and Wu, Wenshan and Mao, Shaoguang and Zhang, Yifan and Tian, Haochen and Vuli{\'c}, Ivan and Zhang, Zhang and Wang, Liang and Tan, Tieniu and others},
  journal={arXiv preprint arXiv:2504.15037},
  year={2025}
}

@misc{zhang2026vibe-benchmark,
  title={How Well Do Models Follow Visual Instructions? VIBE: A Systematic Benchmark for Visual Instruction-Driven Image Editing}, 
      author={Huanyu Zhang and Xuehai Bai and Chengzu Li and Chen Liang and Haochen Tian and Haodong Li and Ruichuan An and Yifan Zhang and Anna Korhonen and Zhang Zhang and Liang Wang and Tieniu Tan},
      year={2026},
      eprint={2602.01851},
      archivePrefix={arXiv},
      primaryClass={cs.CV},
      url={https://arxiv.org/abs/2602.01851}, 
}

@inproceedings{lyu2024unibind,
  title={Unibind: Llm-augmented unified and balanced representation space to bind them all},
  author={Lyu, Yuanhuiyi and Zheng, Xu and Zhou, Jiazhou and Wang, Lin},
  booktitle={Proceedings of the IEEE/CVF Conference on Computer Vision and Pattern Recognition},
  pages={26752--26762},
  year={2024}
}

@inproceedings{li20251+,
  title={Why $1+ 1 < 1$ in Visual Token Pruning: Beyond Naive Integration via Multi-Objective Balanced Covering},
  author={Li, Yangfu and Zhan, Hongjian and Chen, Tianyi and Liu, Qi and Xiong, Yu-Jie and Lu, Yue},
  booktitle={The Thirty-ninth Annual Conference on Neural Information Processing Systems},
  year={2025}
}

@inproceedings{zheng2024learning,
  title={Learning modality-agnostic representation for semantic segmentation from any modalities},
  author={Zheng, Xu and Lyu, Yuanhuiyi and Wang, Lin},
  booktitle={European Conference on Computer Vision},
  pages={146--165},
  year={2024},
  organization={Springer}
}

@inproceedings{jiangdimer,
  title={DiMeR: Disentangled Mesh Reconstruction Model with Normal-only Geometry Training},
  author={Jiang, Lutao and Lin, Jiantao and Chen, Kanghao and Ge, Wenhang and Yang, Xin and Jiang, Yifan and Lyu, Yuanhuiyi and Zheng, Xu and JING, LI and Li, Yinchuan and others},
  booktitle={The Fourteenth International Conference on Learning Representations},
  year={2025}
}
\nocite{zhang2025latent,zhang2025scaling,zhang2026vibe-benchmark,lyu2024unibind,li20251+,lideepscan，lyu2025realrag,zheng2024learning,zheng2024centering，zheng2023both,jiangdimer}

\end{document}